# Selected Qualitative Spatio-temporal Calculi Developed for Constraint Reasoning: A Review

*Debasis Mitra[1]*

*Abstract:* In this article a few of the qualitative spatio-temporal knowledge representation techniques developed by the constraint reasoning community within artificial intelligence are reviewed. The objective is to provide a broad exposure to any other interested group who may utilize these representations. The author has a particular interest in applying these calculi (in a broad sense) in topological data analysis, as these schemes are highly qualitative in nature.

*Index Terms*: temporal knowledge representation, spatial knowledge representation, constraint reasoning, topological data analysis.

## I. INTRODUCTION

Constraint reasoning is an area within artificial intelligence that may be represented as a graph $G=(V, E)$, with nodes in $V$ over some domain $D$ and binary edges in $E$ between those nodes that "constrains" the values that the relevant pairs of nodes can take. A decision problem is to check the satisfiability of constraints, i.e., does there exist a set of values over all nodes such that they locally satisfy all binary constraints in $E$ simultaneously. A qualitative spatio-temporal reasoning (QSTR) problem is defined over a domain $D$ that is space or time $R^n$, for some dimension $n$. [Ligozat, 2012]

In the last four decades many QSTR problems have been studied, primarily for their computational complexities. However, the purpose of this review is not to get into the main constraint reasoning problems studied within the scope of QSTR, but rather to provide an exposure to some of the formalisms (or calculi) developed for the corresponding reasoning problems. Those calculi may have other applications than their intended original purpose. This review does not cover all spatio-temporal calculi (STC) but only a few sample ones that may have the maximum potential in the emerging area of topological data analysis (TDA) within machine learning. [Epstein et al., 2011]

## II. INTERVAL CALCULS

Most of the STC have originated to express qualitative relationships used within the natural languages. Interval Calculus (IC) was developed to relate events each with a finite duration. [Allen, 1983] For example, one may say, *I saw the ambulance pass by during the lecture.* Here, the "lecture" and "ambulance-passing-by" are two events and "during" is a qualitative relationship between the two. As mentioned before, the objective within QSTR was to find out truthfulness of many such statements posted together, as in a detective story. Note that there is no reference to any quantitative value over time in this example. STC only uses qualitative relationships between temporal or spatial entities.

IC is a well-developed formalism that contains exactly 13 jointly exhaustive and pairwise disjoint basic qualitative relationships feasible between any pair of events (Table 1).

| Table 1: Basic IC relations | | | |
|---|---|---|---|
| A ─────    B ───── | A –(b)→ B | B—(a)→A | |
| A ─────   B ───── | A –(m)→ B | B –(m-inv)→A | |
| A ─────   B ───── | A –(o)→ B | B –(o-inv)→A | |
| A ─────   B ───── | A –(f-inv)→ B | B –(f)→A | |
| A ─────   B ───── | A –(d-inv)→ B | B –(d)→A | |
| A ─────   B ───── | A –(s-inv)→ B | B –(s)→A | |
| A ─────   B ───── | A ←(eq)→ B | [Self-converse] | |

Here, b=before, m=meets, o=overlaps, f=finishes, d=during, s=starts, and eq=equals. In each case a suffix *inv* indicates inverse or converse relationship, e.g., *b-inv* means *after*. *eq* is a bi-directional or self-converse relation.

Note that the referred interval (say, *B*) will be at the region *eq* in this space as a point, and the referring interval (say, *A*) will lie at some location depending on its relationship to *B*. Thus, A—(d)→B as in the above example will indicate *A* as a point to be within anywhere in the triangle for *d* near the origin in this space, with respect to *B* located at the point *eq*. A set of intervals could be located in this space when their pairwise relationships are expressed with qualitative IC basic-

_________________
D. Mitra, is with the Computer Science Department, Florida Institute of Technology, 150 West University Blvd, Melbourne, FL, 32901. (Phone: 321-674-8763, Email: dmitra@fit.edu). This article is written while the author was a Visiting Professor in Mathematics at Stanford University.

relations. However, no quantitative information is needed for the purpose.

A time-interval can be represented as a point in a two-dimensional Cartesian space where starting points of intervals are on the horizontal-axis and ending points are on the vertical-axis. [Ligozat, 1996] Of course, any finite interval will be located above the diagonal half-plane passing through the origin in this space. In this *Canonical* representation these thirteen relations partition the space (Figure 1).

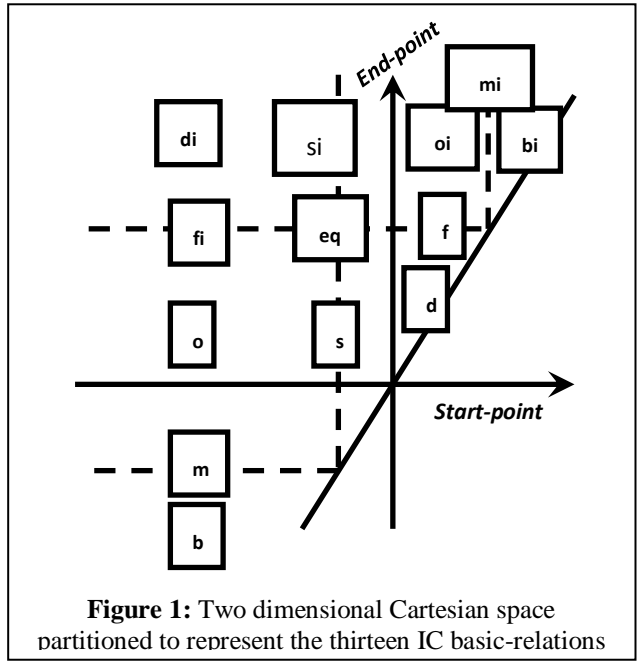

**Figure 1:** Two dimensional Cartesian space partitioned to represent the thirteen IC basic-relations

Shape of an object (e.g., data point clouds) may be represented with a few intervals in a metric space where each interval represents persistence of a shape feature. [Zomorodian et al., 2004; Carlsson, 2009] In a persistence diagram (PD) such intervals are represented in a very similar manner as in IC canonical representation (Figure 1). However, a PD maintains its metric values of start and end points of intervals. I propose that a PD may be represented with IC basic relationships only. [Barannikov, 1994]

A point to mention here for historical completeness is that a time-point calculus also exists in the QSTR literature with three basic relations $\{<, >, =\}$ between two time points on a real line *R*. This calculus is called *point calculus* (PC). [Vilain et al., 1989]

### III. STAR CALCULUI

Star calculi (SC) are extensions of Cardinal-directions calculus over directional relations people use, again in natural languages. Cardinal-directions are based on geographical North, and expresses eight other directions (Figure 2). [Ligozat, 1998]

One can do further refinement on this Cardinal-direction calculus by going into arbitrary angular granularity instead of $90^0$ granularity used in the former. This refinement constitutes a Star-calculus depending on the number of basic relations in it. Thus, Star-9 calculus with 9 basic relations (Figure 2) is the Cardinal-directions calculus, and Star-17 calculus with 17 basic relations is with $45^0$ granularity. Thus, each calculus of the Star-calculi is with a specific angular precision. For example, a $30^0$ precision Star-calculus will have 23 directions, plus the *eq* relation at the center, total 24 basic-relations (Figure 3). [Renz et al., 2004]

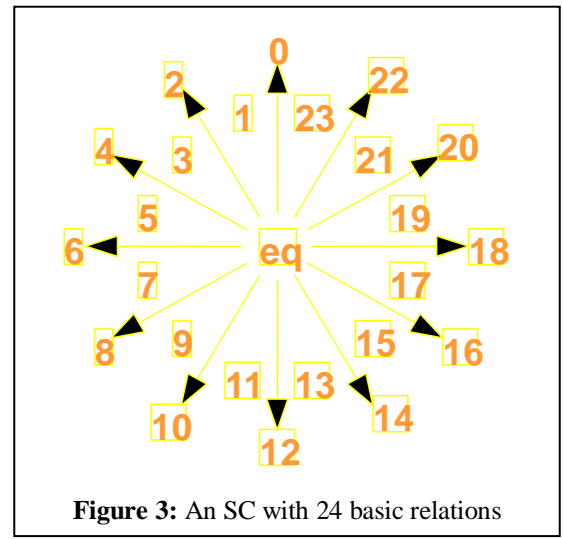

**Figure 3:** An SC with 24 basic relations

Many one-dimensional curves in 2D metric space may be qualitatively represented as a sequence of such basic relations between consecutive pairs of points over the curve where those points are some type of fiducial or representative points.

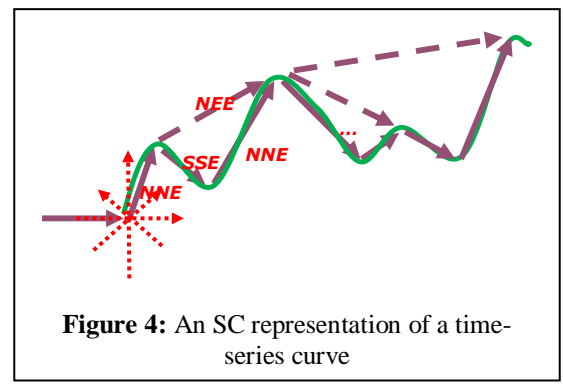

**Figure 4:** An SC representation of a time-series curve

As in Figure 4, fiducial points are selected on a time-series where the signs of slopes reversed (or optimum points on a continuous curve). A directional relation between those points can subsequently represent the shape of the curve qualitatively. [Mitra, 2002]

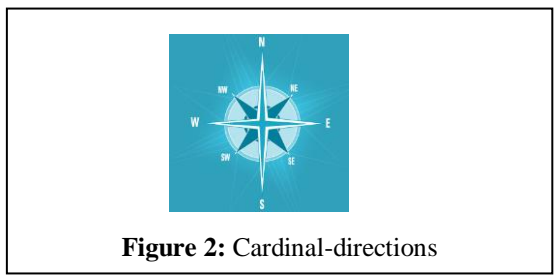

**Figure 2:** Cardinal-directions

When the underlying space is more than two-dimensional one will need an *n*-dimensional SC (for *n*>2). For example, in 3D the basic relations will be over two angular dimensions, polar and azimuthal angles. [Kim et al., 2004]

## IV CYCLIC INTERVAL CALCULUS

Possibly the most exotic of QSTR formalisms is the *cyclic-interval calculus* (CC) over the circumference of a circle where its radius is of no significance. [Osmani, 1999; Balbiani et al., 2000] Our daily clock-values follow such a space. The calculus is defined over intervals on the directed circumference. It appears very similar to the IC mentioned above, but because of the lack of a linear order of points the start-point and end-point has no significance. This causes some IC basic relations to merge and some others to split in the CC. For example, *b* and *bi* in IC merges to only one basic relation that is *disjoint* in CC. However, *o* in IC splits into another type of overlap that is on both sides of the two related intervals. One may call this new basic-relation as *double-overlap*.

Total number of basic relations in CC are sixteen (Table 2). A representation similar to that for IC in Figure 1 for CC has been proposed by Ligozat [Ligozat, 2012] where the start and end points' axes cycle back to themselves. Thus, the canonical representation for CC maps on a cylindrical surface with *b* and *bi* merging together.

Application of this calculus may be in representing topological shape in angle-valued data in TDA. [Burgelea et al., 2015]. I feel this calculus raises many interesting questions in topology as it did in the QSTR literature. Learning on such a modulo space may be important as well for many recurrent neural networks.

**Table 2: Basic relations of Cyclic-interval Calculus**

| | | |
|---|---|---|
| 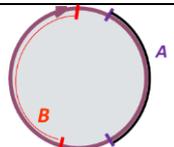 | A ←(disjoint)→ B | [Self-converse] |
| 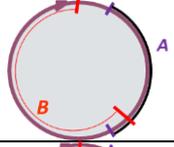 | A —(overlaps)→ B | B —(overlap-inv)→ A |
| 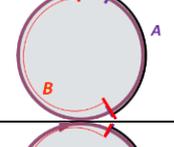 | A —(meets)→ B | B —(meet-inv)→ A |
| 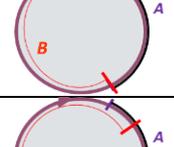 | A ←(double-meets)→ B | [Self-converse] |
| 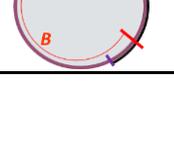 | A ←(double-overlaps)→ B | [Self-converse] |
| 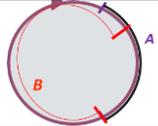 | A --(meets-and-overlapped-by)→ B | B --(met-by-and-overlaps)→ A |
| 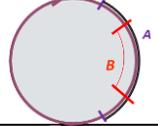 | A --(contains)→ B | B --(contain-inv)→ A |
| 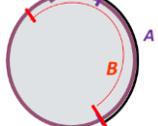 | A --(finishes)→ B | B --(finish-inv)→ A |
| 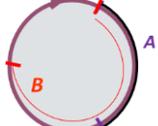 | A --(starts)→ B | B --(start-inv)→ A |
| 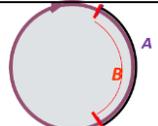 | A ←(equals)→ B | [Self-converse] |

Sixteen basic relations with some of them being by-directional indicating self-converse

## V. REGION-CONNECTION CALACULI

Two types of *region-connection calculi* (RCC) have been invented over two-dimensional metric space, namely, the RCC-5 and RCC-8 where the suffix integer indicates the number of basic relations in the respective calculus. RCC-5 practically contains five jointly exhaustive and pairwise disjoint basic relations possible between two closed sets: *contains, contained, overlaps, disjoint* and *equals*. [Randell et al., 1992] One can say, on a metric space they depict relationships between Venn diagrams between two sets (Figure 5). From this point of view RCC-5 may represent qualitative or topological relationship between open sets. RCC-8 has eight basic relationships where the boundary of a "region" is distinguished from its inside. For this reason, *contains* (and its inverse *inside*) splits into two different basic relations each, e.g., *contains-touching* and *properly-contains* depending on whether the boundaries of two related regions overlap or not. Similarly, *disjoint* may be *completely-disjoint* and *disjoint-touching*. This makes total eight basic relations in

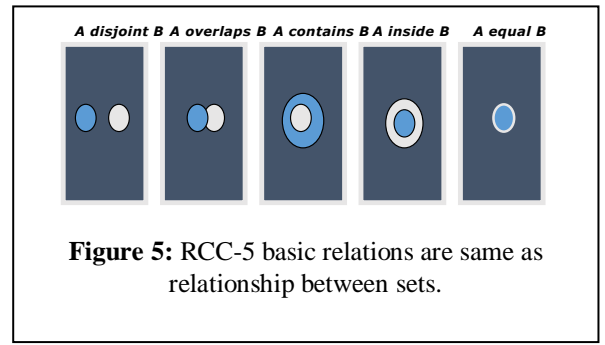

**Figure 5:** RCC-5 basic relations are same as relationship between sets.

RCC-8. [Renz, 2002]

## V. CONCLUSION

Here I have provided eight examples of STC: PC, IC, SC, CC, RCC-5 and RCC-8. Many other STC have been developed in the QSTR literature. My hope is that some of these calculi or their variations will find applications in TDA and other areas. In the past we applied a variation of IC in representing operating system-call sequences to detect anomaly that may represent a malicious attack on a computer. [Tandon et al., 2005] In future we will try to show applications of some of these STC mentioned here to TDA.